\documentclass{article}

\pdfoutput=1

\usepackage{spconf,amsmath,graphicx}
\usepackage{amsmath}
\usepackage{amssymb}
\usepackage{graphicx}
\usepackage{subfig}
\usepackage{tabularx}
\usepackage{booktabs}
\usepackage{microtype}
\usepackage{tikz,pgfplots}
\usepackage{tikz-3dplot}
\usepackage[T1]{fontenc}
\usepackage{hyperref}

\renewcommand{\vec}[1]{\mathbf{#1}}
\def\argmin{\mathop{\mathrm{argmin}}}
\def\argmax{\mathop{\mathrm{argmax}}}

\title{Super-Resolved Retinal Image Mosaicing}
\name{Thomas K\"ohler$^{1,2}$, Axel Heinrich$^1$, Andreas Maier$^{1,2}$, Joachim Hornegger$^{1,2}$, and Ralf P. Tornow$^3$}
\address{$^1$ Pattern Recognition Lab, Friedrich-Alexander-Universit\"at Erlangen-N\"urnberg, Germany\\
$^2$ Graduate School in Advanced Optical Technologies (SAOT), Erlangen, Germany\\
$^3$ Department of Ophthalmology, Friedrich-Alexander-Universit\"at Erlangen-N\"urnberg, Germany
\thanks{The authors gratefully acknowledge funding of the Erlangen Graduate School in Advanced Optical Technologies (SAOT) by the German Research Foundation (DFG) in the framework of the German excellence initiative.
}}
\begin{document}
\ninept
\maketitle
\begin{abstract}
The acquisition of high-resolution retinal fundus images with a large field of view (FOV) is challenging due to technological, physiological and economic reasons. This paper proposes a fully automatic framework to reconstruct retinal images of high spatial resolution and increased FOV from multiple low-resolution images captured with non-mydriatic, mobile and video-capable but low-cost cameras. Within the scope of one examination, we scan different regions on the retina by exploiting eye motion conducted by a patient guidance. Appropriate views for our mosaicing method are selected based on optic disk tracking to trace eye movements. For each view, one super-resolved image is reconstructed by fusion of multiple video frames. Finally, all super-resolved views are registered to a common reference using a novel polynomial registration scheme and combined by means of image mosaicing. We evaluated our framework for a mobile and low-cost video fundus camera. In our experiments, we reconstructed retinal images of up to $30^\circ$ FOV from $10$ complementary views of $15^\circ$ FOV. An evaluation of the mosaics by human experts as well as a quantitative comparison to conventional color fundus images encourage the clinical usability of our framework.
\end{abstract}

\section{Introduction}

Over the past years, digital imaging technologies have been established in ophthalmology to examine the human retina in an in-vivo and non-invasive way \cite{Patton2006}. Digital fundus cameras and scanning laser ophthalmoscopes (SLO) are some of the most commonly used systems to capture single images or video sequences of the retina \cite{Abramoff2010}. This is an essential part for the diagnosis of retinal diseases, e.\,g. in computer-assisted glaucoma \cite{Koehler2015} or diabetic retinopathy screening \cite{Zhang2014}. In intraoperative applications, the slit lamp is a common technique for live examination of the eye background \cite{Richa2014}. Common to all of these approaches is the strong need for capturing high-resolution images with a wide field of view (FOV) to employ them for diagnosis or intervention planning. However, in retinal imaging this is difficult due to technological and economic reasons. First, for a wide FOV the pupil should be dilated. Moreover, the spatial resolution is limited by the characteristics of the camera sensor and optics. Modern fundus cameras and SLO are able to provide images with sufficient resolution to support diagnostic procedures but they are relatively expensive and not mobile limiting their benefits for low-cost screening applications.
 
For these reasons, two complementary software-based strategies are an emerging field of research. (i) Image mosaicing to register and combine multiple views showing different regions of the retina has been proposed to increase the FOV. Can et al. \cite{Can2002} and later Zheng et al. \cite{Zheng2014a} have developed feature-based registration schemes based on vascular landmark matching that are applicable to mosaicing of high-resolution fundus images acquired longitudinally. Similarly, interest points can be used for feature-based registration \cite{Cattin2006}. In more recent approaches, intensity-based \cite{Adal2014} or hybrid image registration \cite{Richa2014,Chanwimaluang2006} have been proposed to avoid the need for accurate feature extraction. Common to these methods is that they either rely on high-quality data or they are applicable to images of poor quality but cannot enhance them, e.\,g. in terms of resolution. (ii) For spatial resolution enhancement of digital images, super-resolution has been investigated. In recent works from K\"ohler et al. \cite{Koehler2014} and Thapa et al. \cite{Thapa2014}, multiple low-resolution frames from a video sequence showing the same FOV but with small geometric displacements to each other are fused to reconstruct a high-resolution image. This approach exploits complementary information across different frames due to small natural eye motion during an examination. Unlike image mosaicing, the super-resolution paradigm cannot be employed to increase the FOV as all frames have to cover the same region. Early work in computer vision \cite{Zomet2000} suggests the combination of both, using super-resolution reconstruction applied to a mosaic.

This paper proposes a novel multi-stage framework to reconstruct super-resolved retinal mosaic images. Unlike many related approaches, we exploit video sequences rather than longitudinally acquired images. As a key idea, we use a mobile and video-capable but low-cost camera to scan different regions on the retina as typically done, e.\,g. using fundus video cameras or slit lamps. We propose eye tracking to select appropriate views for our reconstruction method in a fully automatic way. Complementary to related concepts \cite{Richa2014,Zomet2000}, these are first fused by super-resolution reconstruction followed by image mosaicing. For accurate combination of the views, we introduce robust intensity-based registration and a novel adaptive weighting scheme. Our experimental evaluation performed with a low-cost fundus camera demonstrates the clinical practicality of our method.

\section{Super-Resolved Mosaicing Framework}

\begin{figure*}[!t]
	\centering
	\includegraphics[width=0.88\textwidth]{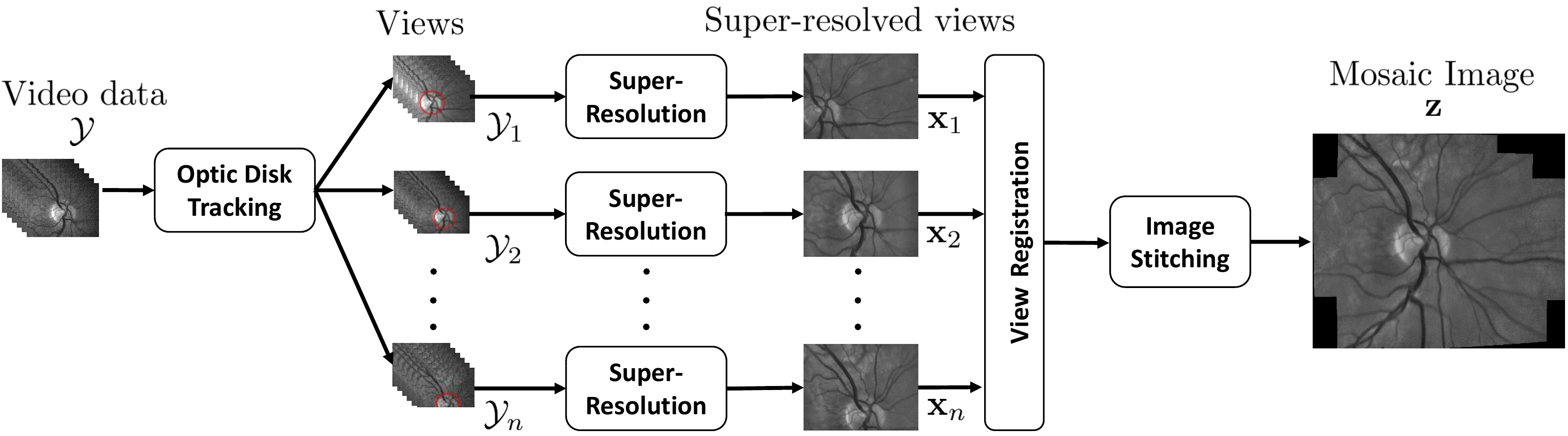}
	\caption{Pipeline of the proposed multi-stage framework for super-resolved mosaicing from retinal video sequences.}
	\label{fig:pipeline}
\end{figure*} 

We consider a video sequence of $K$ frames denoted by the set $\mathcal{Y} = \{ \vec{y}^{(1)}, \ldots, \vec{y}^{(K)} \}$, where each frame $\vec{y}^{(k)} \in \mathbb{R}^M$ is represented in vector notation. The frames in $\mathcal{Y}$ show different regions of the retina, whereas each region is captured by $K_i$  frames $\mathcal{Y}_i = \{ \vec{y}_i ^{(1)}, \ldots, \vec{y}_i ^{(K_i)} \}$ and $\mathcal{Y}_i$ is referred to as a \textsl{view}. To apply our mosaicing framework within the scope of one examination, we scan different regions on the retina by exploiting eye movements conducted by a patient guidance. Our approach aims at reconstructing a super-resolved mosaic in a three-stage procedure as depicted in Fig.~\ref{fig:pipeline}. (i) In order to select appropriate views for super-resolved mosaicing, we employ eye tracking to trace the eye position during the examination. (ii) For $n$ views $\mathcal{Y}_1, \ldots, \mathcal{Y}_n$ automatically selected by the eye tracking, we reconstruct high-resolution images $\mathcal{X} = \{ \vec{x}_1, \ldots, \vec{x}_n \}$ by means of multi-frame super-resolution where $\vec{x}_i \in \mathbb{R}^N$, $N > M$. (iii) Finally, the complementary views in $\mathcal{X}$ are registered to a common reference and are combined by image stitching.

\subsection{Super-Resolution View Reconstruction}

Appropriate views for mosaicing are selected based on eye tracking in an initial stage. For this purpose, we employ the geometry-based tracking-by-detection introduced by K{\"u}rten et al. \cite{kurten2014}. This enables real-time tracking of the optic disk as a robust feature to describe eye motion. For each frame $\vec{y}^{(k)}$, the tracking yields the optic disk radius and the pixel coordinates of its center point \smash{$\vec{u}_{\text{eye}}^{(k)}$}. Based on the coordinates \smash{$\vec{u}_{\text{eye}}^{(k)}$}, we decompose the entire input sequence $\mathcal{Y}$ into $n$ disjoint subsets $\mathcal{Y}_i$. For each view $\mathcal{Y}_i$, we compute the euclidean distance \smash{$d(\vec{u}_i, \vec{u}_{\text{eye}}^{(k)})$} for consecutive positions relative to $\vec{u}_i$ describing the eye position in the first frame of $\mathcal{Y}_i$. Each view is composed of $K_i$ consecutive frames as 
\smash{$\mathcal{Y}_i = \{ \vec{y}^{(k)} : d(\vec{u}_{i}, \vec{u}_{\text{eye}}^{(k)}) \leq d_{\text{max}} \}$},
where $d_{\text{max}}$ is the maximum amount of motion accepted within one view. The starting positions $\vec{u}_{i}$ are selected such that $d(\vec{u}_{i-1}, \vec{u}_{i}) \geq d_{\text{min}}$, where $d_{\text{min}}$ is the minimum distance between two successive views to gain an improvement in terms of FOV. The view $\mathcal{Y}_r$ with the closest distance $d(\vec{u}_r, \vec{u}_0)$ to the centroid $\vec{u}_0$ of all views is selected as reference to ensure that $\mathcal{Y}_r$ has sufficient overlap to all other views. 

For each view $\mathcal{Y}_i$, we obtain a super-resolution reconstruction based on the frames corresponding to this view. This reconstruction exploits subpixel motion in $\mathcal{Y}_i$ that is related to small, natural eye movements that occur during an examination. We adopt the adaptive algorithm presented in our prior work \cite{Koehler2014} and estimate $\vec{x}_i$ via:
\begin{equation}
	\begin{split}
		\vec{x}_i  = \argmin_{\vec{x}} &  \sum_{k = 1}^{K_i} \left| \left| \vec{y}_i^{(k)} - \boldsymbol{\gamma}_{m,i}^{(k)} \odot \vec{W}_i^{(k)}\big(\boldsymbol{\theta}_i^{(k)} \big) \vec{x} - \gamma_{a,i}^{(k)} \vec{1} \right| \right|_1\\ 
		&+ \lambda(\vec{x}) \cdot R(\vec{x}),
	\end{split}
	\label{eqn:srObjectiveFunction}
\end{equation}
where the term $R(\vec{x})$ denotes bilateral total variation (BTV) regularization and $\lambda(\vec{x}) \geq 0$ denotes an adaptive regularization weight. The parameter vector \smash{$\boldsymbol{\theta}_i^{(k)}$} encodes subpixel motion to describe eye movements by an affine transformation and \smash{$\vec{W}_i^{(k)} \in \mathbb{R}^{KM \times N}$} is the system matrix to model $\smash{\boldsymbol{\theta}_i^{(k)}}$, sub-sampling and the camera point spread function. The multiplicative and additive parameters \smash{$\boldsymbol{\gamma}_{m,i}^{(k)}$} and \smash{$\gamma_{a,i}^{(k)}$} model spatial and temporal illumination changes that are estimated by bias field correction. The regularization weight $\lambda(\vec{x})$ is adaptively selected using image quality self-assessment. To achieve uniform noise and sharpness characteristics across all views and hence consistency required for mosaicing, this parameter is initially determined for the reference view according to: 
\begin{equation}
	\lambda_r = \argmax_{\lambda} Q\big\{ \vec{x}_r(\lambda) \big\},
\end{equation}
where $Q\{\vec{x}_r(\lambda)\}$ denotes the quality measure to assess the appearance of $\vec{x}_r(\lambda)$ that is reconstructed using the weight $\lambda$ \cite{Koehler2014}. Once $\lambda_r$ is determined, Eq.\,\eqref{eqn:srObjectiveFunction} is solved for each view $\vec{x}_i$ with a fixed regularization weight using scaled conjugate gradient iterations.

\begin{figure*}[!t]
	\centering
	\setlength{\tabcolsep}{1.2pt}
	\begin{minipage}[t][][b]{0.43\textwidth}
		\begin{tabular}{ccc}
			\includegraphics[width=0.32\textwidth]{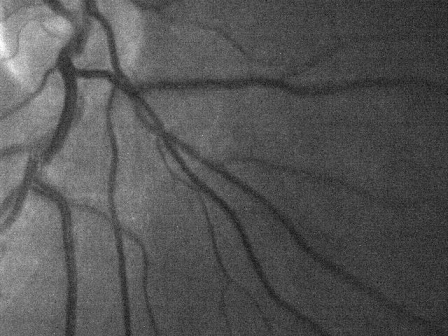}
			& \includegraphics[width=0.32\textwidth]{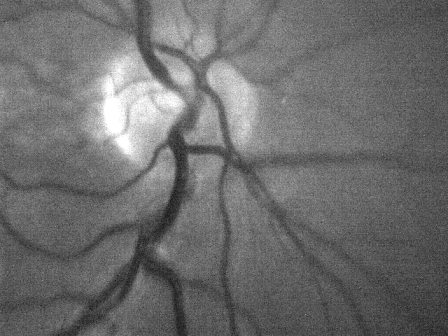}
			& \includegraphics[width=0.32\textwidth]{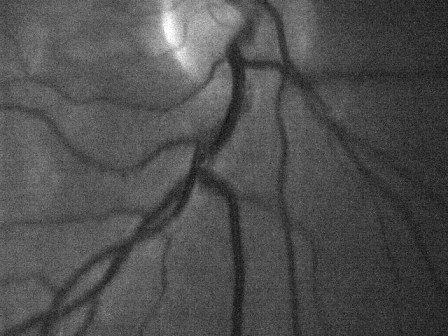}\\
			\includegraphics[width=0.32\textwidth]{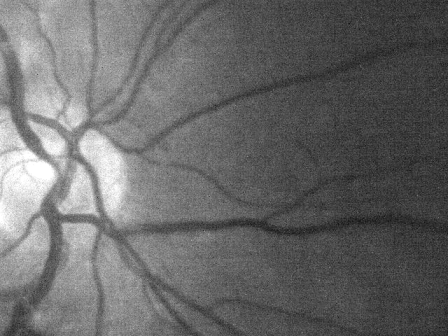}
			& \includegraphics[width=0.32\textwidth]{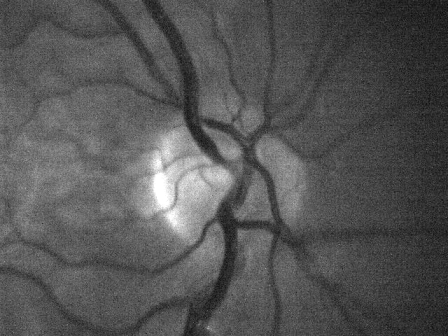}
			& \includegraphics[width=0.32\textwidth]{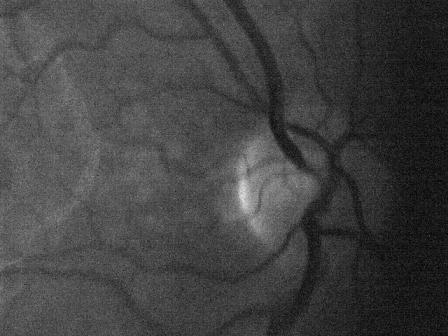}\\
			\includegraphics[width=0.32\textwidth]{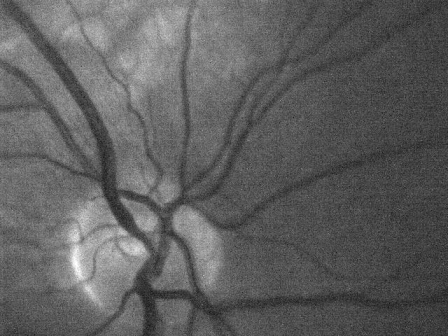}
			& \includegraphics[width=0.32\textwidth]{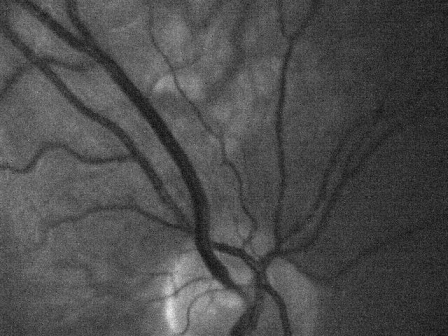}
			& \includegraphics[width=0.32\textwidth]{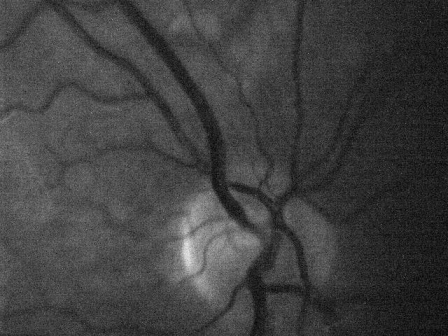}\\[12.2ex]
			\includegraphics[width=0.32\textwidth]{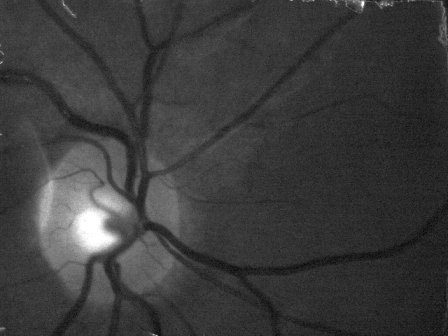}
			& \includegraphics[width=0.32\textwidth]{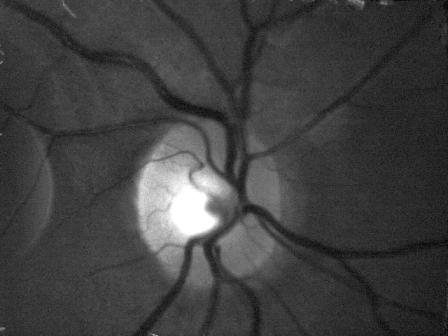}
			& \includegraphics[width=0.32\textwidth]{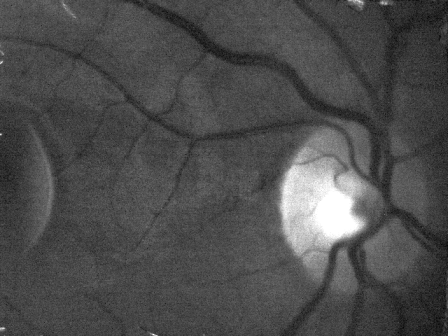}
		\end{tabular}
	\end{minipage}
	\qquad
	\begin{minipage}[t][][b]{0.43\textwidth}
		\centering
		\begin{tabular}{c}
			\includegraphics[width=0.881\textwidth]{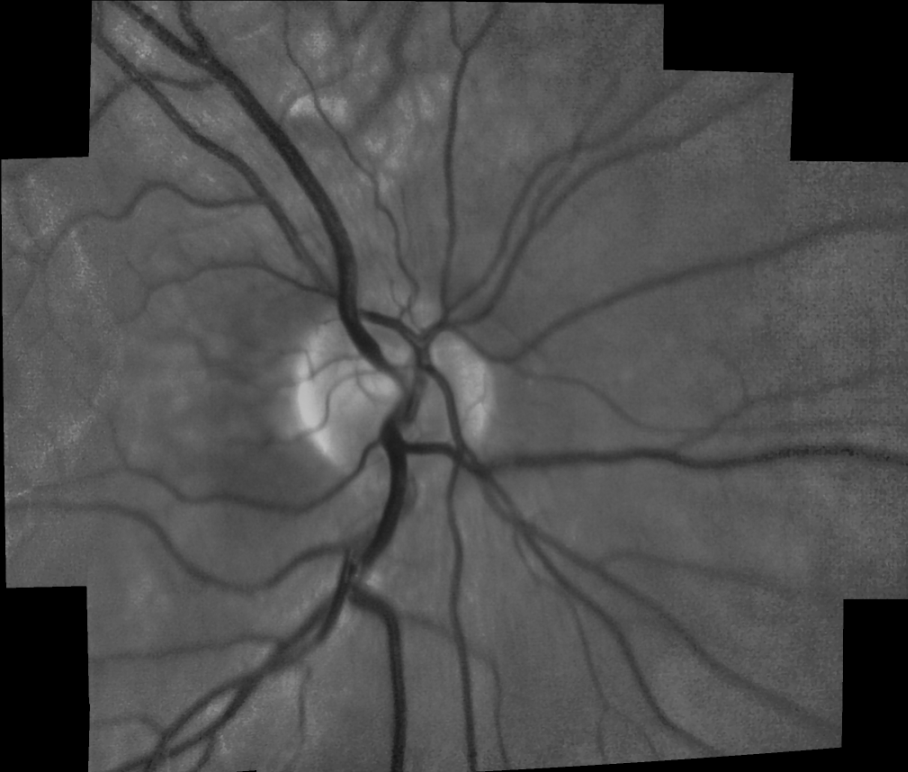}\\[0.5ex]
			\includegraphics[width=0.881\textwidth]{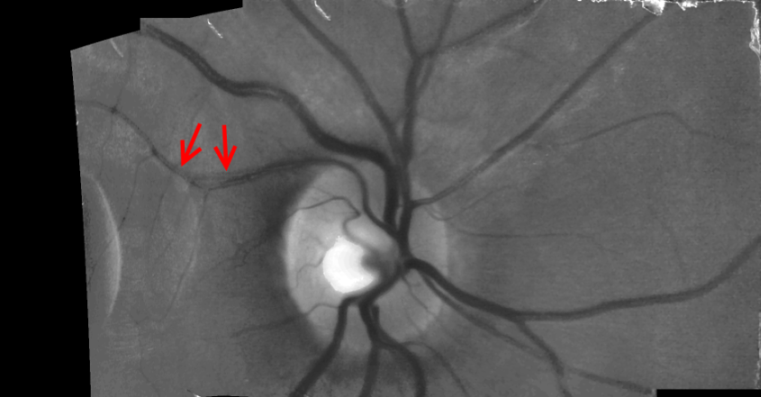}
		\end{tabular}
	\end{minipage}
	\caption{Left: video data acquired with a mobile low-cost camera. Right: super-resolved mosaics obtained from $n = 10$ views with quality grade \smash{'perfect'} (top row) and $n = 3$ views with quality grade \smash{'poor'} due to locally inaccurate registration (bottom row).}
	\label{fig:SRImages}
\end{figure*}

\subsection{Mosaic Image Reconstruction}

We propose a fixed-reference registration scheme for robust mosaicing that is insensitive to error accumulation. The super-resolved views $\vec{x}_i$, $i \neq r$ are registered to the reference $\vec{x}_r$ as selected by the automatic tracking procedure. For view registration, we employ a 12 degrees of freedom (DoF) quadratic transformation to consider the spherical surface of the retina \cite{Can2002}. A point $\vec{u} = (u_1, u_2)^\top$ in $\vec{x}_i$ is transformed to $\vec{u}^\prime = Q(\vec{u}, \mathbf{p})$ in $\vec{x}_r$ according to:
\begin{equation}
	\setlength{\arraycolsep}{1.0pt}
	\vec{u}^\prime
	= 
	\begin{pmatrix}
		p_1 &\ p_2 &\ p_3 &\ p_4 &\ p_5 &\ p_6\\
		p_7 &\ p_8 &\ p_9 &\ p_{10} &\ p_{11} &\ p_{12} 
	\end{pmatrix}
	\begin{pmatrix}
		u_1^2 &\ u_2^2 &\ u_1u_2 &\ u_1 &\ u_2 &\ 1 
	\end{pmatrix}^\top,
	\label{eqn:transmat}
\end{equation}
where $\vec{p} \in \mathbb{R}^{12}$ denotes the transformation parameters. To apply this model for view registration, we propose intensity-based registration. This approach does not rely on accurate feature detection, e.\,g. vascular tree segmentation, that is hard to achieve in retinal images of lower quality. As photometric differences between multiple views are an additional issue, we adopt the correlation coefficient as a similarity measure $\rho: \mathbb{R}^N \times \mathbb{R}^N \rightarrow [0; 1]$. This measure has been investigated to estimate projective transformations using an enhanced correlation coefficient (ECC) optimization algorithm \cite{Evangelidis2008} to maximize $\rho(\vec{x}_i, \vec{x}_r)$ iteratively. Iterations are performed according to $\vec{p}^{t} = \vec{p}^{t-1} + \vec{\Delta p}(J(\vec{u},\vec{p}))$, where $\vec{\Delta p}(J(\vec{u},\vec{p}))$ is the increment for the parameters $\vec{p}$ at iteration $t$ computed from a scaled version of the Jacobian of $Q(\vec{u},\mathbf{p})$. The proposed method generalizes this framework to the quadratic model in Eq.\,\eqref{eqn:transmat}, where the Jacobian of $Q(\vec{u},\mathbf{p})$ with respect to $\mathbf{p}$ is computed per pixel $\vec{u}$ as:
\setcounter{MaxMatrixCols}{12}
\begin{equation}
	\setlength{\arraycolsep}{1pt}
	J(\vec{u},\vec{p}) = 
	\begin{pmatrix}
		u_1^2 &\ u_2^2 &\ u_1u_2 &\ u_1 &\ u_2 &\ 1 &\ 0 &\ 0 &\ 0 &\ 0 &\ 0 &\ 0\\
		0 &\ 0 &\ 0 &\ 0 &\ 0 &\ 0 &\ u_1^2 &\ u_2^2 &\ u_1u_2 &\ u_1 &\ u_2 &\ 1
	\end{pmatrix}
\end{equation}
The registration of view $\vec{x}_i$ to the reference $\vec{x}_r$ is implemented in a hierarchical scheme to avoid getting stuck in a local optimum. We employ the eye positions $\vec{u}_i$ and  $\vec{u}_r$ obtained from the tracking procedure to estimate a translational motion $\vec{t}_i = \vec{u}_i - \vec{u}_r$ and initialize our model in Eq.\,\eqref{eqn:transmat} by $\vec{P}_{\text{trans}} = (\vec{0}_{2 \times 5}, \vec{t}_i)$. Based on this initialization, we estimate the model $\vec{P}_{\text{affine}} = (\vec{0}_{2 \times 3}, \vec{A})$ equivalent to a 6 DoF affine model defined by $\vec{A} \in \mathbb{R}^{2 \times 3}$. Finally, we estimate the full quadratic model $\vec{P}_{\text{quad}} \in \mathbb{R}^{2 \times 6}$ in Eq.\,\eqref{eqn:transmat} using the affine model as initial guess.
In addition to the geometric registration, mosaicing requires photometric registration to compensate for illumination differences across the views. Therefore, histogram matching is employed to determine a monotonic mapping that adjusts the histogram of each view $\vec{x}_i$, $i \neq r$ to the histogram of the reference $\vec{x}_r$. 

Applying geometric and photometric registration to $\vec{x}_i$ yields the registered view $\tilde{\vec{x}}_i$, whereas for $i = r$ we set $\tilde{\vec{x}}_r = \vec{x}_r$. 
To reconstruct a mosaic $\vec{z}$ from $\tilde{\vec{x}}_1, \ldots, \tilde{\vec{x}}_n$ by image stitching, we propose the pixel-wise adaptive averaging \cite{capel2004}: 
\begin{equation}
	\vec{z}(\vec{u}) = \frac{1}{\sum_{i = 1}^{n(\vec{u})} \vec{w}_i(\vec{u})} \sum_{i = 1}^n \vec{w}_i(\vec{u}) \tilde{\vec{x}}_i(\vec{u}),
\end{equation}
where $\vec{w}_i$ are adaptive weights. For reliable mosaicing, the weights need to be selected such that seams between overlapping views are suppressed. Moreover, robust mosaicing needs to account for the registration uncertainty of individual views. In our approach, these issues are addressed by the adaptive weights:
\begin{equation}
	\vec{w}_i(\vec{u}) = 
	\begin{cases}
		\pi_i(\vec{u}) \kappa(\vec{u}) \rho(\vec{v}_i, \vec{v}_r) 	& \text{if } \rho(\vec{v}_i, \vec{v}_r) > \rho_{v,\text{min}} \\
																																																								& \,\, \wedge\,\,\rho(\vec{x}_i, \vec{x}_r) > \rho_{i,\text{min}} \\
		0																																				& \text{otherwise}
	\end{cases},
\end{equation}
where $\pi_i: \mathbb{R}^{N \times N} \rightarrow \{0, 1\}$ is an indicator function with $\pi_i(\vec{u}) = 1$ if the $i$-th view contributes to the mosaic at position $\vec{u}$ and $\pi_i(\vec{u}) = 0$ otherwise. The spatially varying weights $\kappa(\vec{u})$ are computed by a distance map of $\tilde{\vec{x}}_i$ that decays from a maximum at the image center to zero at the boundary. $\rho(\cdot, \cdot)$ denotes the correlation evaluated on the vesselness filtered images \cite{Frangi1998} $\vec{v}_i$ and $\vec{v}_r$ as well as on the intensity images $\vec{x}_i$ and $\vec{x}_r$ to assess the registration uncertainty in overlapping views. To exclude incorrectly registered views from mosaicing based on their consistency with the reference, $\rho_{v,\text{min}} \in [0; 1]$ and $\rho_{i,\text{min}} \in [0; 1]$ are pre-defined thresholds for the correlation values.

\begin{figure*}[!t]
  \centering
	\ninept
	\begin{minipage}[b]{0.46\textwidth}
	\centering
		\ninept
		\begin{tabular}{p{8.0em} rrr}
			\toprule\noalign{\smallskip}
			\textbf{Quality grade} & \textbf{Exp.\,1} & \textbf{Exp.\,2} & \textbf{Exp.\,3} \\
			\midrule
			Perfect (1)			&	8					& 18					& 3 \\
			\midrule
			Acceptable (2)		& 12			& 6					& 17 \\
			\midrule
			Poor (3)					& 4				& 0				& 4 \\
			\midrule
			Not usable (4)		& 0			& 0				& 0 \\
			\bottomrule
		\end{tabular}
	\end{minipage}
	\quad
	\begin{minipage}[b]{0.21\textwidth}
		\begin{tabular}{c}
			\includegraphics[width=0.94\textwidth]{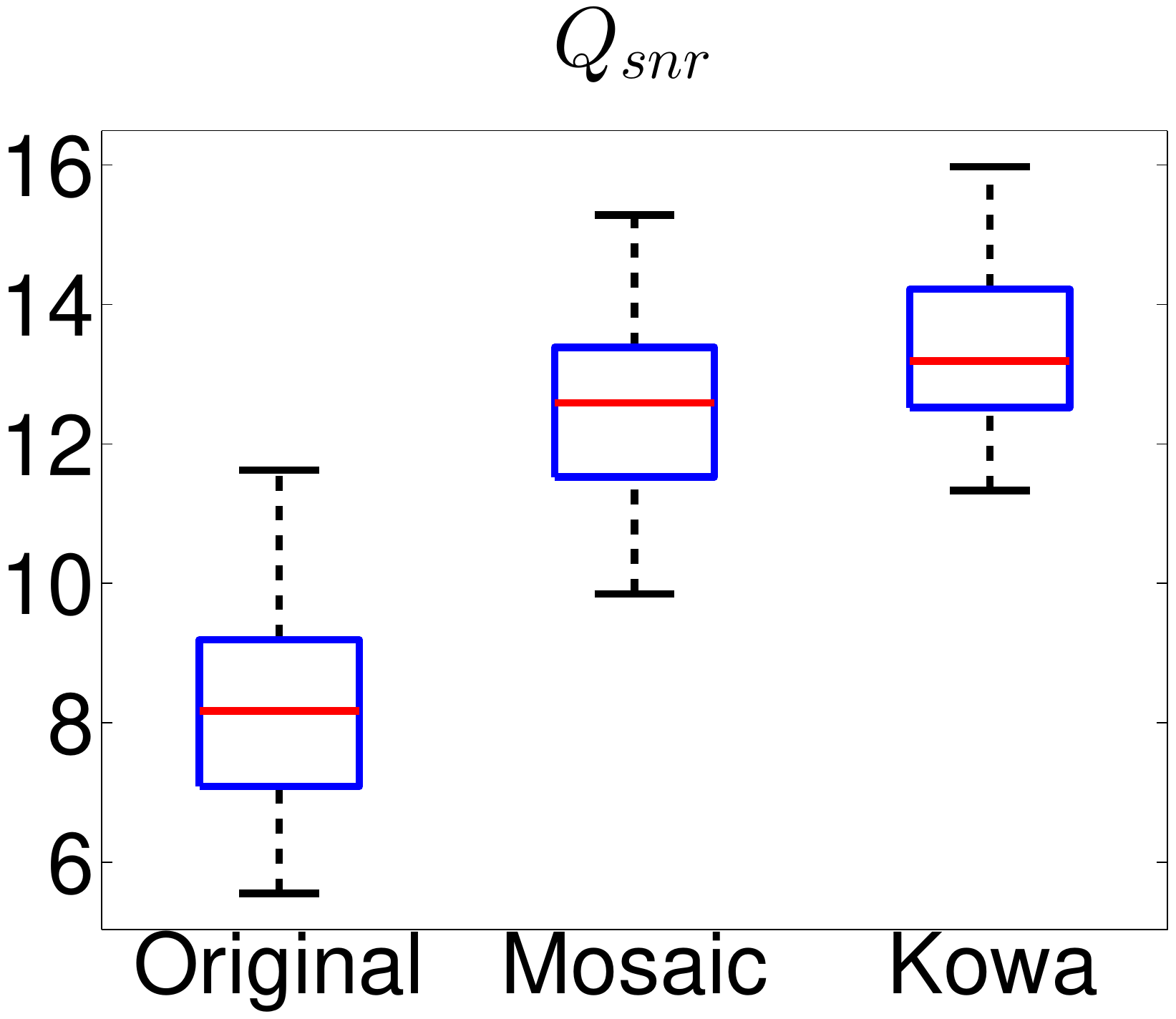}
		\end{tabular}		
  \end{minipage}
	\quad
	\begin{minipage}[b]{0.21\textwidth}
		\begin{tabular}{c}
			\includegraphics[width=0.94\textwidth]{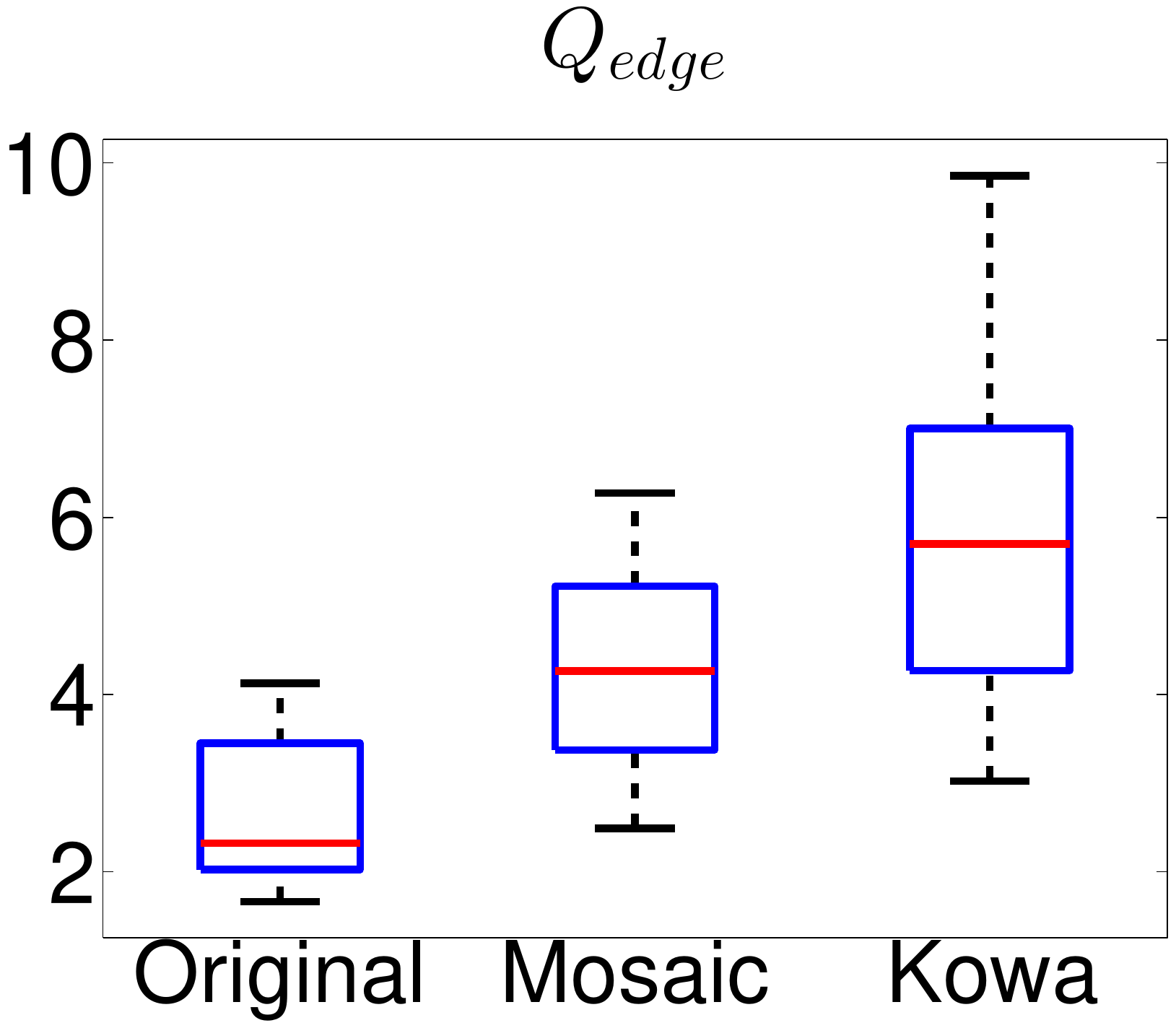}
		\end{tabular}		
  \end{minipage}
	\hfill
	\caption{Left: quality grades provided by three human experts for 24 super-resolved mosaics. Right: boxplots of blind signal-to-noise ratio $Q_{\text{snr}}$ and edge preservation $Q_{\text{edge}}$ for original video data, super-resolved mosaics and high-resolution images acquired with a Kowa nonmyd camera.}
  \label{fig:qualityScores}
\end{figure*}

\section{Experiments and Results}

\begin{figure*}[!t]
	\centering
	\begin{tabular}{cc}
		\includegraphics[width=0.41\textwidth]{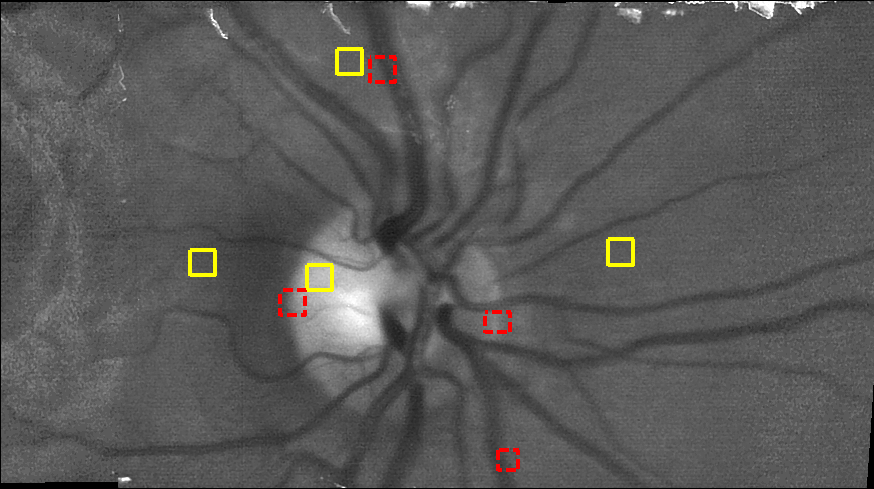} 
		& \quad \includegraphics[width=0.302\textwidth]{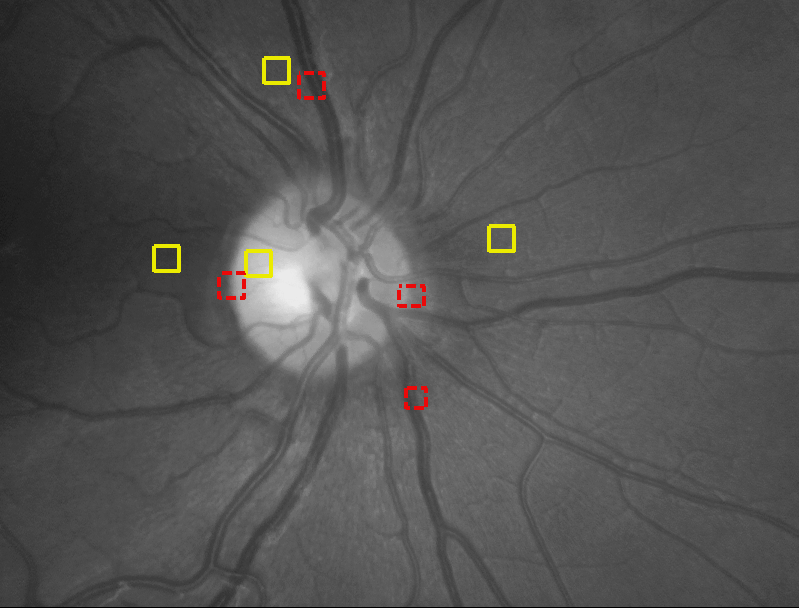}
	\end{tabular}
	\caption{Left: mosaic graded as 'acceptable'. Right: grayscale converted image of the same region acquired with a commercial Kowa nonmyd camera. The ROIs that were used to compute $Q_{\text{snr}}$ and $Q_{\text{edge}}$ for quantitative comparisons are highlighted in yellow and red, respectively.}
	\label{fig:compare}
\end{figure*}

We demonstrate the application of our framework for fundus imaging using the mobile and low-cost camera presented in \cite{Tornow2015} to acquire monochromatic images with a frame rate of 25\,Hz. We examined the left eye of seven healthy subjects with a FOV of $15^\circ$ in vertical and $20^\circ$ in horizontal direction without pupil dilation. To scan different regions on the retina, we fixed the camera position and asked the subjects to fixate different positions on a fixation target. The duration of each video was $\approx$ 15\,s and images were given in VGA resolution ($640 \times 480$\,px). In total, we acquired 24 data sets. We aligned the camera such that the optic disk was centered in the first frame, which was selected as the reference view. We varied the number of views between $n = 2$ and $n = 10$. The view selection was performed with $d_{\text{max}} = 5$\,px, $d_{\text{min}} = 100$\,px and $K_i = 6$. We employ our public available super-resolution toolbox\footnote{The latest version of our toolbox is available on our webpage www5.cs.fau.de/research/software/multi-frame-super-resolution-toolbox/} to reconstruct super-resolved views with  $2 \times$ magnification and apply mosaicing with $\rho_{i,\text{min}} = 0.5$ and $\rho_{v,\text{min}} = 0.1$.
The super-resolved mosaics for 24 datasets were assessed by three human experts in retinal imaging. Each image was ranked in the following categories: (i) Quality of the geometric registration and appearance of anatomical structures. (ii) Homogeneity of the illumination on the retina. (iii) Overall appearance of the reconstructed image. Each category was graded ranging from 'perfect' (grade: 1) to 'not usable' (grade: 4). The overall grade for each image was chosen to be the worst of the three categories. Two example images obtained with different grades are shown in Fig.\,\ref{fig:SRImages}. These were obtained by horizontal and vertical eye movements and the FOV was increased up to $\approx 30^\circ$. The overall distribution of the quality grades is summarized in Fig.\,\ref{fig:qualityScores} (left). 

To compare our approach to cameras used in clinical practice, we captured color fundus images with a Kowa nonmyd camera ($25^\circ$\,FOV, $1600 \times 1216$\,px). Fig.\,\ref{fig:compare} compares a mosaic with a grayscale converted fundus image captured from the same subject, where mosaicing enhanced the horizontal FOV to $\approx 25^\circ$ based on $n = 3$ views. This was comparable to the FOV provided by the Kowa camera. For quantitative evaluation, we examined the blind signal-to-noise ratio $Q_{\text{snr}}$ and edge preservation $Q_{\text{edge}}$ measured by:
\begin{align}
	Q_{\text{snr}}(\vec{x}) &= 10\,\log_{10} \left( \mu_{\text{flat}} / \sigma_{\text{flat}} \right) \\
	Q_{\text{edge}}(\vec{x}) &= \frac{w_b (\mu_b - \mu)^2 + w_f (\mu_f - \mu)^2}{w_b \sigma_b^2 - w_f\sigma_f^2}.
\end{align}
Here, $\mu_{\text{flat}}$ and $\sigma_{\text{flat}}$ are the mean and standard deviation of the intensity within a homogenous region of interest (ROI) in $\vec{x}$. Similarly, $w_i$, $\mu_i$ and $\sigma_i$ with $i \in \{b, f\}$ denote the weight, the mean and the standard deviation of a Gaussian mixture model fitted for background (b) and foreground (f) in an ROI containing a transition between two structures and $\mu$ is the mean intensity in this ROI. Fig.\,\ref{fig:qualityScores} (right) compares the statistics of $Q_{\text{snr}}$ and $Q_{\text{edge}}$ of original video data, super-resolved mosaics and the Kowa images for five subjects using boxplots. For both measures, we evaluated four manually selected ROIs per image.

In our experiments, $89$\,\% of the mosaics were ranked as 'acceptable' or 'perfect' without noticeable artifacts and severe registration errors were alleviated by our adaptive mosaicing scheme, see Fig.\,\ref{fig:SRImages} (top). No image was graded as 'not usable' and $11$\,\% were graded as 'poor' due to a low contrast of videos for individual subjects, which is not enhanced by our framework. In the remaining cases, images were graded as 'poor' due to inaccurate geometric registrations in individual regions, see Fig.\,\ref{fig:SRImages} (bottom).
In these experiments with a non-mydriatic camera, which makes imaging with wide FOV challenging, the proposed framework was able to provide a spatial resolution and FOV comparable to those of high-end cameras, see Fig.\,\ref{fig:compare}. In particular, our method was able to double the FOV compared to the original video. In terms of \smash{$Q_{\text{snr}}$} and \smash{$Q_{\text{edge}}$}, we obtained substantial improvements by super-resolved mosaicing compared to original video data and competitive results compared to the Kowa camera. 
Unlike related methods, the benefit of our framework is that mosaicing is applicable even with mobile and low-cost video hardware within one examination of a few seconds rather than longitudinal examinations. This is essential in computed-aided screening, where our approach provides a relevant alternative to expensive and non-mobile cameras as typically used in clinical practice.

\section{Conclusion and Future Work}

In this work, we have proposed a fully automatic framework to reconstruct high-resolution retinal images with wide FOV from low-resolution video data showing complementary regions on the retina. Our approach exploits super-resolution to obtain multiple super-resolved views that are stitched to a common mosaic using intensity-based registration with a quadratic transformation model. Using a mobile and low-cost video camera, our framework is able to reconstruct retinal mosaics that are comparable to photographs of commercially available high-end cameras in terms of resolution and FOV.

One scope of future work is mosaicing of peripheral retinal areas as our current framework processes central areas around the optic nerve head. Another promising direction for future research is the formulation of super-resolution and mosaicing in a joint optimization approach to further enhance the robustness of mosaic reconstruction.

\bibliographystyle{IEEEbib}
\ninept
\bibliography{bibliography}

\end{document}